\pdfoutput=1

\documentclass[11pt]{article}
\usepackage{graphicx}
\usepackage{multirow}
\usepackage{subcaption}
\usepackage{ACL2023}
\usepackage{tabularx}
\graphicspath{ {./images/} }

\usepackage{times}
\usepackage{latexsym}
\usepackage{booktabs}
\usepackage[rightcaption]{sidecap}

\usepackage{wrapfig}

\usepackage[T1]{fontenc}

\usepackage[utf8]{inputenc}

\usepackage{microtype}

\title{IUST\textunderscore NLP at SemEval-2023 Task 10:
Explainable Detecting Sexism with Transformers and Task-adaptive Pretraining}

\author{Hadiseh Mahmoudi \\
  School of Computer Engineering \\
  Iran University of Science and Technology   \\
  \texttt{Hadisehmahmoudy@gmail.com} \\\
  }
\begin{document}
\maketitle
\begin{abstract}
This paper describes our system on SemEval-2023 Task 10: \emph{Explainable Detection of Online Sexism (EDOS).} This work aims to design an automatic system for detecting and classifying sexist content in online spaces. We propose a set of transformer-based pre-trained models with task-adaptive pretraining and ensemble learning. The main contributions of our system include analyzing the performance of different transformer-based pre-trained models and combining these models, as well as providing an efficient method using large amounts of unlabeled data for model adaptive pretraining. We have also explored several other strategies. On the test dataset, our system achieves F1-scores of 83\%, 64\%, and 47\% on subtasks A, B, and C, respectively.
\end{abstract}
\section{Introduction}
Discriminatory views against women in online environments can be extremely harmful so in recent years it has become a serious problem in social networks. Identifying online sexism involves many challenges because sexist discrimination and misogyny have different types and appear in different forms. Therefore, the aim of the EDOS shared task \cite{kirkSemEval2023} is to develop English-language models for detecting sexism. These models should be more accurate and explainable and include detailed classifications for sexist content from Gab and Reddit. 
 This task covers a wide range of sexist content and aims to distinguish different types of sexist content. There are three hierarchical subtasks within the task:

\begin{itemize}
\setlength\itemsep{0.5cm}
\item \textbf{SubTask A: Binary Sexism Detection.} This subtask involves a binary classification problem, where the goal is to determine whether a given post is sexist or non-sexist using predictive systems.
\item \textbf{SubTask B: Category of Sexism.} When classifying sexist posts, a four-class classification system is used, where the predictive systems are tasked with identifying one of the following categories: (1) threats, (2) derogation, (3) animosity, or (4) prejudiced discussions.
\item \textbf{SubTask C: Fine-grained Vector of Sexism.} When dealing with sexist posts, a fine-grained classification approach is utilized, where predictive systems are required to assign one of 11 specific vectors to each post.
\end{itemize}
\par In this paper, we describe our system for three subtasks. First, we used some models based on transformers. These models are fine-tuned for classification and in addition, they are used by adding classification components in the upper layer. We have also created a model by combining several transformer-based models and adding a classification layer and using it in all three subtasks as ensemble learning. We then built a task-adaptive model to adapt it to our specific domain. We trained this model on a large unlabeled dataset and fine-tuned it on labeled data for all subtasks. For all three sub-tasks, we used cross-entropy loss for multi-class classification.
Our best system was able to achieve an F1-score of 83\% (subtask A), 64\% (subtask B), and 47\% (subtask C) on the test dataset.

\section{Background}
\begin{figure}[t]
    \centering
    \includegraphics[width=6cm, height=5cm]{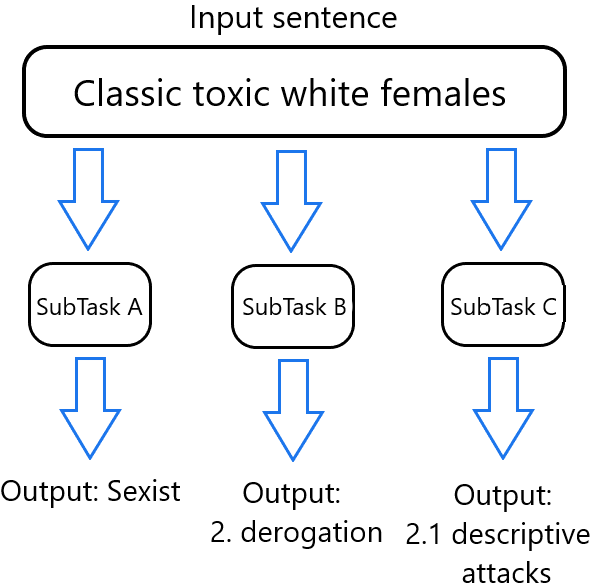}
    \caption{Sample input and output for sexist content. The samples that have not-sexist labels in subtask A have no labels in subtask B and subtask C.}
    \label{fig:sample_input_output}
\end{figure}
Detecting and addressing sexist content in online spaces has emerged as a significant issue. In this work, we leverage a dataset obtained from two online networks, Reddit and Gab, comprising of user-generated text content. While 20,000 of the dataset are labeled, a substantial portion of it (two million) remains unlabeled. Our system employs both datasets to perform the subtasks. Figure \ref{fig:sample_input_output} displays a sample input and output for each subtask.

\par State-of-the-art results in Natural Language Processing (NLP) tasks have been achieved by pre-trained models such as BERT, RoBERTa, ERNIE, XLM-RoBERTa, and DeBERTa \cite{bert2018,robeta,ernie, xlmRoberta, deberta}. These models, which are trained on large datasets with numerous parameters, have demonstrated the potential to improve system performance and generalization in recent studies. Therefore, we used several pre-trained models such as BERT \cite{bert2018}, RoBERTa \cite{robeta}, XLM-RoBERTa \cite{xlmRoberta}, competitive pre-trained models such as DeBERTa \cite{deberta}, and large-scale pre-trained models such as ERNIE \cite{ernie}. While pre-trained models have demonstrated excellent performance in sentence-level or paragraph-level tasks and binary classification subtask A, our preliminary tests revealed that they did not perform well in subtasks B and C, which involve classifying into four and eleven classes, respectively. The limited amount of training data available, differences in the scope of our work, and pre-trained models' training can explain these results. However, training these models in our specific domain can potentially enhance the results.
\par In recent years, there has been significant research into detecting sexism on social media platforms. Several approaches have been proposed, ranging from rule-based models to machine learning algorithms. Binary classification of shared content in online networks as sexist and not-sexist has been considered in many recent works and Some of the best methods used in sexism detection include deep learning, ensemble methods, and hybrid approaches. In this work \cite{work1}, a new work is proposed that aims to understand and analyze how sexism is expressed, from overt hate or violence to subtle expressions, in online conversations. the authors first collected a dataset of 10,000 tweets on various topics including politics, sports, culture, etc. Then, using this dataset, they trained different models for sexism detection using deep learning methods. For example, they used convolutional neural networks, recurrent neural networks, and hybrid neural networks. The performance of each model was then evaluated using accuracy, precision, and recall metrics. Finally, the authors concluded that convolutional and hybrid neural networks are more accurate and precise in detecting sexism on Twitter. In \cite{work2} binary and multi-class classification, two multilingual transformer models are used, one based on multilingual BERT and the other based on XLM-R. This paper's approach uses two different strategies to adapt transformers to detect sexist content. first, unsupervised pre-training with additional data, and second, supervised fine-tuning with additional and augmented data. The authors compare their model's performance with several state-of-the-art methods, including traditional machine learning models and deep learning models like convolutional neural networks and long short-term memory networks. The evaluation results show that the proposed multilingual transformer model outperforms all other methods, achieving state-of-the-art performance in detecting sexism across multiple languages.
\par The structure of our paper is as follows: Section \ref{S3} describes our methodological approach and describes the models used. Our experimental setup and employed datasets will be described in Section \ref{S4} of this paper, followed by a documentation of the results (Section \ref{S5}) and a final discussion and conclusion (Section \ref{S6}).
\section{System Overview}
\label{S3}
This section describes the methods we have used in our system.
The main challenge in this work is the rather small number of training datasets for all three subtasks (subtask A includes 14,000 training instances, and subtasks B and C include 3,398 training instances). This rather small size makes it difficult to effectively pre-training complex NLP methods such as transformers. For this reason, we use different transfer learning methods. As a basis for textual content modeling, we apply several pre-trained transformer-based models: BERT, DeBERTa, and RoBERTa. To adapt these general-purpose models to the task of recognition (subtask A) and sexism classification (subtask B and C), we use a large unlabeled dataset that has the same content as the training set. Also, in another approach, we combine several pre-trained models and train them by adding a classification layer for each subtask.
\par In Section \ref{ss3.1} our system that uses transformer-based models is described for three subtasks. Then, we describe our task-adaptive system on the domain of this particular task (see Section \ref{ss3.2}). Also, the ensemble model system is described in Section \ref{ss3.3}. Finally, we describe the optimization algorithm and loss function in Section \ref{ss3.4}.
\begin{figure}[t]
    \centering
    \includegraphics[width=6cm, height=5cm]{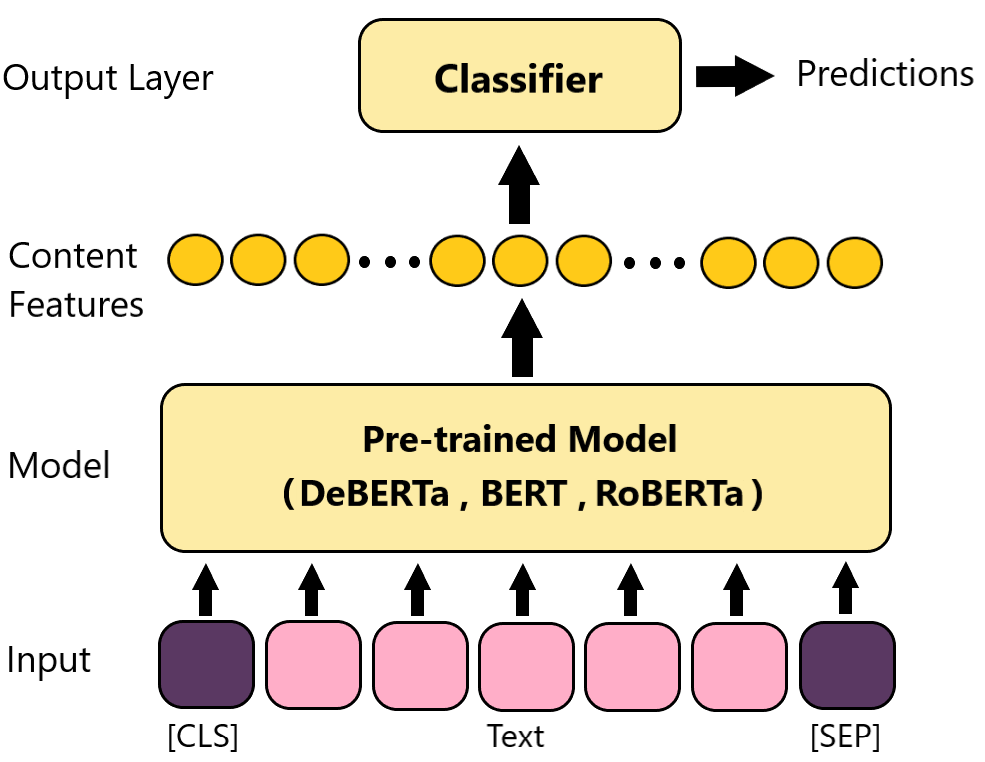}
    \caption{The architecture of the pre-trained transformer-based model. In this architecture, various pre-trained models such as BERT, DeBERTa, and RoBERTa are used. The output layer is different for each subtask. Subtask A includes two outputs, subtask B includes 4 outputs, and subtask C includes 11 outputs.}
    \label{fig:transformer_models}
\end{figure}
\subsection{Transformer-based Model}
\label{ss3.1}
The architecture of this model is shown in Figure \ref{fig:transformer_models}. For all three subtasks A, B, and C, we experimented with pre-trained transformer-based models such as BERT, RoBERTa, and DeBERTa. Since the distribution of classes in all three subtasks is unbalanced, model training is conducted with and without class-weights\footnote{This argument allows you to define float values to the importance to apply to each class.}in the loss function. Also, the model is fine-tuned using the training dataset. The output layer of the model is different for each subtask, there are two classes for subtask A, four classes for subtask B, and 11 classes for subtask C. The classification layer of the pre-trained transformer-based model for sequence classification was utilized.

\subsection{Task-adaptive Model}
\label{ss3.2}
\begin{figure}[t]
    \centering
    \includegraphics[width=4cm, height=6cm]{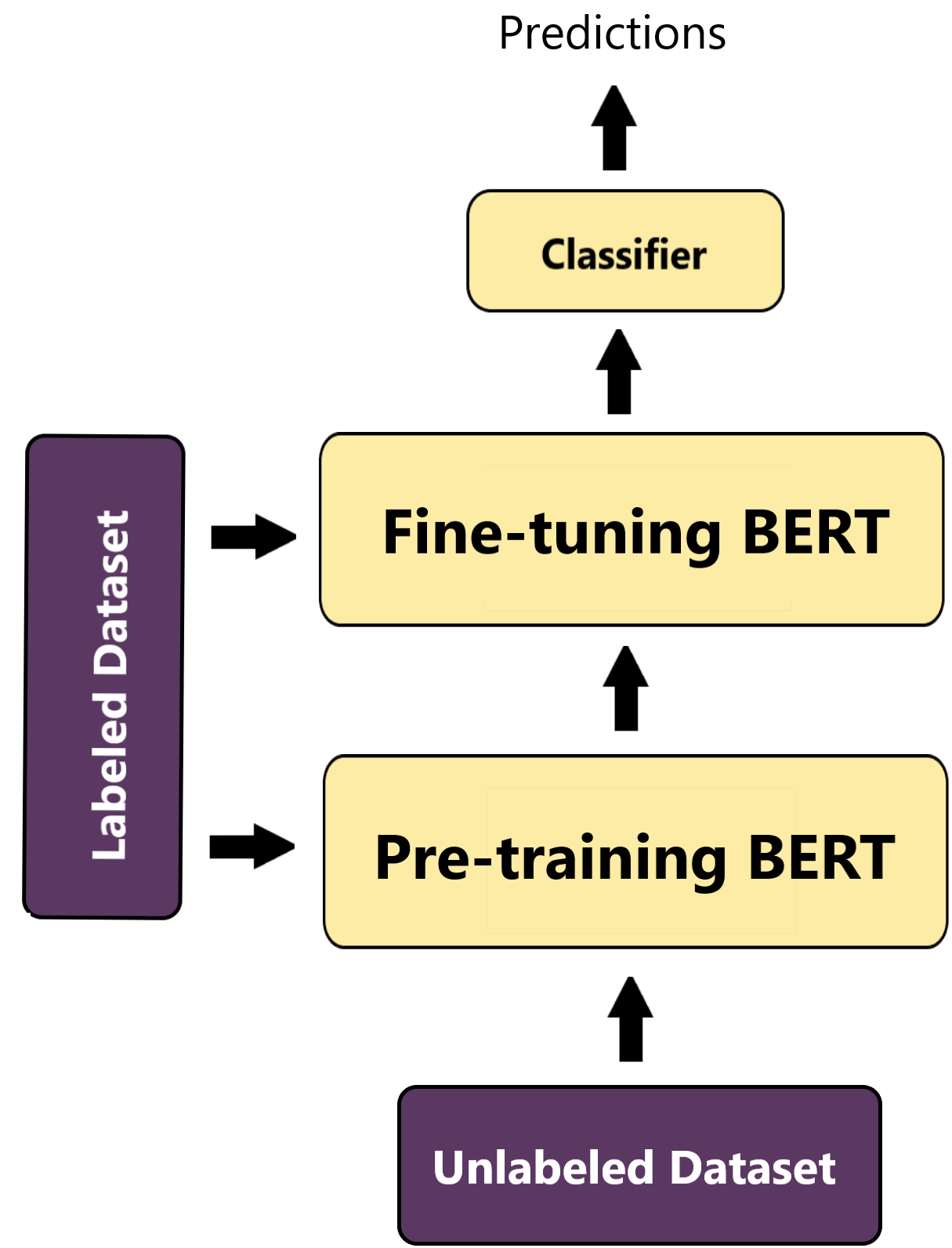}
    \caption{The architecture of the task-adaptive model. The output layer is different for each subtask. Subtask A includes two outputs, subtask B includes 4 outputs, and subtask C includes 11 outputs.}
    \label{fig: pretraining}
\end{figure}
Transformer-based models can overfit small data sets \cite{overfit}. Therefore, we used pre-trained models for all subtasks to overcome this issue. Furthermore to adapt our model to the scope of this particular task, in our experiments, we trained the pre-trained transformer model in an unsupervised manner on a large unlabeled dataset. This large unlabeled dataset was provided by \cite{kirkSemEval2023}, and it contains unlabeled examples from the same domain (i.e, also from Gab and Reddit). We only used the pre-trained BERT model. Then we fine-tuned the adaptive model with the labeled data, and finally, we make predictions for each subtask. The architecture of our task-adaptive model can be seen in Figure \ref{fig: pretraining}.

\subsection{Ensemble Model}
\label{ss3.3}
\begin{figure*}[t]
\centering
    \includegraphics[width=0.9\textwidth , height=8cm]{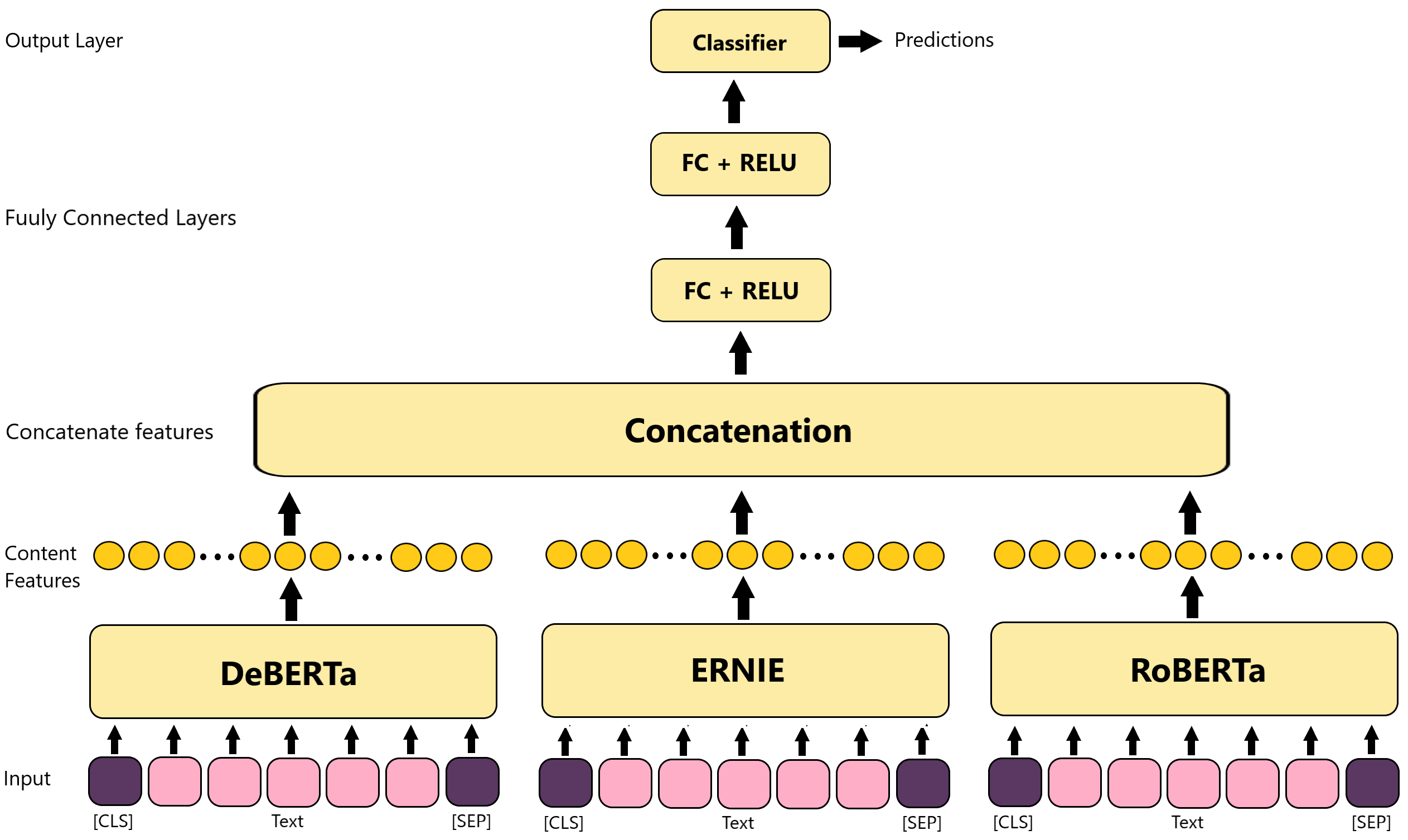}
    \caption{Ensemble learning model architecture. Subtask A includes two outputs, subtask B includes 4 outputs, and subtask C includes 11 outputs.}
    \label{fig: ensemble}
\end{figure*}
In this model, we experiment with the combination of transformer-based models. The architecture of the ensemble model is shown in Figure \ref{fig: ensemble}. The input is given to three pre-trained models, then the output of these models, which are the representation of the input sentence, are combined. In this model, we tested different combinations of transformer-based models. Our ensemble model takes the input and feeds it into three pre-trained models. We then added a dropout layer to the outputs of each pre-trained model. Finally, we combined the outputs of the dropout layers to obtain a representation of the input sentence. A first dense layer with a Relu activation function and a dropout layer as follows:
\begin{equation} 
 h_1 = Dropout(Relu(h_t W_1+b_1))
\end{equation}
where \( h_1 \in R^{1\times d}\), \(W_1 \in R^{d\times d}\) and \(b_1 \in R^d\) are the embedding, learnable weight, and bias of the first dense layer respectively. \( h_t \in R^{3\times 768}\) It is also a combination of the output of pre-trained models. The last dense layer is as follows:
\begin{equation}
 h_{out} = Dropout(Relu(h_1 W_2 + b_2))
\end{equation}
where \( h_{out} \in R^{1\times d}\), \(W_2 \in R^{d\times d}\) and \(b_2 \in R^d\) are the final embedding of the last layer, learnable weight, and bias respectively. Finally, the \(h_{out}\) is transformed to fit the two-class for subtask A, four-class for subtask B and 11-class for subtask C classification.

\subsection{Optimization and Loss Function}
\label{ss3.4}
For all three subtasks A, B, and C, we used the cross-entropy loss function for multi-class classification as follows:
\begin{equation}
 CE = -\Sigma_{c=1}^M y_{i,c} log(p_{i,c})
\end{equation}
where \(M\) denotes the number of classes, \(y_{i,c}\) is a binary indicator (0 or 1) that indicates whether c is a correct class of i-th instance and \(p_{i,c}\) means the predicted probability of class c. 

\section{Experimental Setup}
\label{S4}
\begin{table}[t]
\centering
\begin{tabular}{c c} 
 \hline
 \hline
 \textbf{Hyperparameter} & \textbf{Value} \\
 \hline
 Dropout & 0.5 \\
 \hline
 Learning Rates & 2e-5, 3e-5, 5e-5 \\
 \hline
 Optimizer & AdamW \\
 \hline
 Learning schedule & Step Decay  \\ 
 \hline
 schedule Step size & 3  \\ 
 \hline
 Epochs & 2, 3, 5, 10, 15  \\ 
 \hline
 Batch Size & 16,32,48  \\ 
 \hline  
 \hline  
\end{tabular}
\caption{\label{parameter-table}The values of the hyperparameters for fine-tuning the models that we used in all three subtasks. Due to unbalanced data, we used class-weight in the loss function of some models.}
\end{table}
\begin{table}[t]
\begin{tabular}{|c c c c|} 
 \hline
 \textbf{Task} & \textbf{Train} & \textbf{Dev} & \textbf{Test} \\ 
 \hline
 \hline
 SubTask A & 14000 & 2000 & 4000 \\ 
\hline
 SubTask B & 3398 & 486 & 970 \\
 \hline
 SubTask C & 3398 & 486 & 970 \\
 \hline
  \hline
 \textbf{Unlabeled dataset} & 2 million & - & -  \\
  \hline
\end{tabular}
\caption{\label{dataset-table}A summary of the dataset. Data from subtasks B and C include samples that are labeled as sexist in subtask A.}
\end{table}
To fine-tune the pre-trained transformer-based models used in our systems, we employed the AdamW optimizer \cite{adam} with default hyperparameters. Additionally, we explored the impact of additional hyperparameters, as detailed in Table \ref{parameter-table}. To determine the optimal values of the hyperparameters, we utilized the dev dataset. Experiments were done with Nvidia GTX 1080.

\subsection{Data}
We used the training, validation, and test data sets provided by SemEval2023 Task 10 \cite{kirkSemEval2023}. The statistics of the corpus are presented in Table \ref{dataset-table}. In all three subtasks, 70\% of the total data is allocated to training data, 10\% to dev data, and 20\% to test data. Also, for the task-adaptive system, we used the unlabeled data set provided to us.
\subsection{Evaluation}
Our system makes classifications for all three subtasks. Therefore, the output of the model is predicting one class from two classes for subtask A, predicting one class from four classes for subtask B, and predicting one class from 11 classes for subtask C. To evaluate the results, we used F1-score and other metrics such as precision and recall.

\section{Results}
 \label{S5}
 \begin{table}[t]
\centering
\begin{tabular}{c|c|c}
 \hline
\hline
\textbf{Task} & BERT\(_{base}\) & DeBERTa \\
\hline
\textbf{SubTask A} & 0.7750 & - \\
 \hline
 \textbf{SubTask B} & - & 0.6364   \\
 \hline
 \textbf{SubTask C} & 0.4230  & -  \\
 \hline
 \hline
\end{tabular}
\caption{\label{leaderboard-table}Official results on the leaderboard for three subtasks. These results are reported on test sets.}
\end{table}

 \begin{table*}[t]
 \renewcommand{\arraystretch}{1.2}
\begin{tabularx}{\textwidth}{X |X |X |X}
 \hline
 \hline
   & \hspace{0.9cm}\textbf{SubTask A} & \hspace{0.9cm}\textbf{SubTask B} & \hspace{0.9cm}\textbf{SubTask C} \\
\hline
   \textbf{Model} & \hspace{0.5cm}Dev\hspace{1.1cm}Test & \hspace{0.5cm}Dev\hspace{1.1cm}Test & \hspace{0.5cm}Dev\hspace{1.1cm}Test \\
 \hline
  BERT\(_{base}\) & \hspace{0.3cm}0.8036\hspace{0.7cm}0.7587& \hspace{0.3cm}0.6159\hspace{1.1cm}- & \hspace{0.3cm}0.3975\hspace{1.1cm}- \\
 \hline
  BERT\(_{base}\)+\hspace{0.1cm}Class-Weight & \hspace{0.3cm}0.8257\hspace{0.7cm}0.7834 & \hspace{0.3cm}0.5926\hspace{1.1cm}- & \hspace{0.3cm}0.3597\hspace{1.1cm}- \\
 \hline
 RoBERTa\(_{base}\) & \hspace{0.3cm}0.8120\hspace{1.1cm}- & \hspace{0.3cm}0.6055\hspace{1.1cm}- & \hspace{0.3cm}0.3754\hspace{1.1cm}- \\
 \hline
 DeBERTa\(_{base}\) & \hspace{0.3cm}0.8418\hspace{0.7cm}\textbf{0.8334} & \hspace{0.3cm}\textbf{0.6841}\hspace{0.7cm}0.6035 & \hspace{0.3cm}\textbf{0.4855}\hspace{0.7cm}0.4166 \\
 \hline
 DeBERTa\(_{base}\)+\hspace{0.1cm}Class-Weight & \hspace{0.3cm}\textbf{0.8421}\hspace{0.7cm}0.8319 & \hspace{0.3cm}0.6737\hspace{0.7cm}\textbf{0.6382} & \hspace{0.3cm}0.4672\hspace{0.7cm}\textbf{0.4728} \\
 \hline
 Ensemble (DeBERTa, RoBERTa, ERNIE) & \hspace{0.3cm}0.8336\hspace{0.7cm}0.8326 & \hspace{0.3cm}0.6687\hspace{0.7cm}0.6254 & \hspace{0.3cm}0.3322\hspace{0.7cm}0.3206\\
 \hline
 Task-adaptive BERT\(_{base}\) Model & \hspace{0.3cm}0.8362\hspace{0.7cm}0.8324 & \hspace{0.3cm}0.6545\hspace{0.7cm}0.6112 & \hspace{0.3cm}0.4436\hspace{0.7cm}0.4344  \\
 \hline
 \hline
\end{tabularx}
\caption{\label{results-table}The results of the F1-scores for the described systems and for all three subtasks.}
\end{table*}

\begin{table*}[t]
 \renewcommand{\arraystretch}{1.2}
\begin{tabularx}{\textwidth}{ X  X  X | X | X | X }
 \hline
\hline
   \textbf{Model} & & \textbf{Batch Size} & \hspace{0.3cm}\textbf{SubTask A} & \hspace{0.3cm}\textbf{SubTask B} & \hspace{0.3cm}\textbf{SubTask C} \\
 \hline
  DeBERTa\(_{base}\)+ & & \hspace{0.5cm} 16 & \hspace{0.6cm}0.8179 & \hspace{0.6cm}0.6223 & \hspace{0.6cm}0.4056 \\
 Class-Weight & & \hspace{0.5cm} \textbf{32} & \hspace{0.6cm}\textbf{0.8421} & \hspace{0.6cm}\textbf{0.6737} & \hspace{0.6cm}\textbf{0.4672} \\
 \hline
 \multirow{2}{4em}{Ensemble} & & \hspace{0.5cm} 32 & \hspace{0.6cm}0.8148 & \hspace{0.6cm}0.6312 & \hspace{0.6cm}0.2974 \\ 
& & \hspace{0.5cm} \textbf{48} & \hspace{0.6cm}\textbf{0.8336} & \hspace{0.6cm}\textbf{0.6687} & \hspace{0.6cm}\textbf{0.3322} \\ 
 \hline
Task-adaptive & & \hspace{0.5cm} 16 & \hspace{0.6cm}0.8035 & \hspace{0.6cm}0.6152 & \hspace{0.6cm}0.3991 \\
(BERT\(_{base}\))& & \hspace{0.5cm} \textbf{32} & \hspace{0.6cm}\textbf{0.8362} & \hspace{0.6cm}\textbf{0.6545} & \hspace{0.6cm}\textbf{0.4436} \\
 \hline
 \hline
\end{tabularx}
\caption{\label{results-batch_table}F1-scores of our best models on the dev set for all three subtasks and different batch sizes.}
\end{table*}

\begin{figure}[t]
\centering
    \includegraphics[width=7cm, height=5cm]{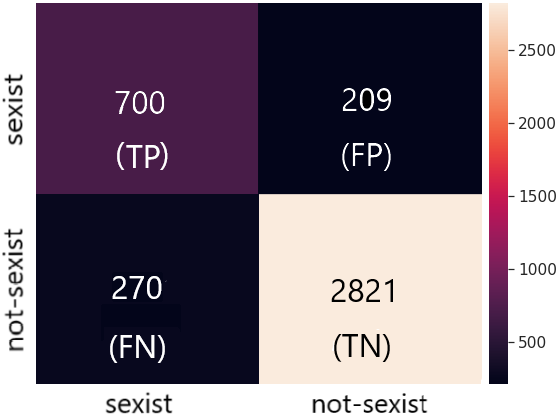}
    \caption{Confusion matrix of DeBERTa model on subtask A without any class-weight. These predictions were made on the dev set. }
    \label{fig:confusion_A}
\end{figure}

In this section, we examine the results of the systems that we explained in the previous section. Our best system was able to Macro-F1 score 83\%, 64\%, and 47\% on the test sets and 84\%, 68\%, and 49\% on dev sets in subtasks A, B, and C, respectively. Using the BERT model, we obtained the official result submitted for subtask A on the test dataset. Our official submission system was based on the BERT model, in which we utilized class-weight in the loss function. We then improved it using other models such as DeBERTa and ensemble models. Also, the official results for subtasks B and C have been using DeBERTa's model, which in this paper we have improved by adding class weights to the cross-entropy loss function. These weights are calculated based on the ratio of the number of samples in each class to the total number of samples. They are then normalized and used as the weight of each class in the loss function. The official scores that we have submitted are as follows: 77\% in subtask A, 64\% in subtask B, and 42\% in subtask C, which can be seen in Table \ref{leaderboard-table}.
\par Table \ref{results-table} shows the results of our different models. These results are based on the Macro-F1 score. The performance benchmark of our systems is the final results of the test dataset, but for better understanding, we have also presented the results of the models on the dev dataset. To obtain the performance of our systems on the test dataset, we first trained the mentioned models with different values of hyperparameters and then choose the optimal hyperparameters on the dev set. Finally, for our system and combination of hyperparameters, we make predictions on the held-out test set. We presented the average results of at least three tests as the final results on the test set. To deal with the unbalanced data, we used the class-weight method in the Loss function, which could improve the results, especially in subtasks B and C. Also, in all models, the learning rate of \(2e-5\) had the best performance on the dev set. The results of our main models on three subtasks' dev sets with different batch sizes are presented in Table \ref{results-batch_table}. In the pre-trained and task-adaptive models, batch size 32 and in the ensemble model batch size 48 gave the best results.
\par Using the DeBERTa\(_{base}\), task-adaptive, and ensemble models, we were able to achieve promising results in subtask A, with the best performance belonging to model DeBERTa\(_{base}\) on the test set. For subtasks B and C, we obtained the best results on the test sets using the DeBERTa\(_{base}\) model with class-weight. The classification layer of the pre-trained DeBERTa\(_{base}\) for sequence classification model was utilized in the DeBERTa\(_{base}\) model. However, in the ensemble model, pre-trained transformer-based models are only used for feature extraction. Then, several dense layers are added to combine the extracted features, and finally, an output layer is added for classification, which is trained on the data. Due to the small number of data samples and their insufficient amount for fine-tuning the model, the ensemble could not achieve better results, especially in subtask C, which had very few data samples and a higher number of classes compared to other subtasks.
\par A confusion matrix of the DeBERTa model predictions on subtask A, is shown in Figure \ref{fig:confusion_A}. Because the classes are unbalanced, the false negative error is more than the false positive error.
 
\section{Conclusion}
\label{S6}
In this work, we described our system for SemEval2023 Task 10. The purpose of this task was to identify and classify sexist content in online spaces. To do this, we used different types of transformer-based models as well as their combinations. We also presented a task-adaptive system in the domain of this particular task. In the above sections, we examined the essential components for ensuring the repeatability of our systems and evaluated the performance of the architectures. Our best system was able to achieve an F1-score of 83\% in subtask A, 64\% in subtask B, and 47\% in subtask C on test sets. In future work, larger pre-trained models such as DeBERTa-v3 can be used for classification or task-adaptative pretraining, and it is also promising to use other unsupervised methods or use more data to train the task-adaptive model.

\section{Acknowledgments}
I am grateful to the organizers of the SemEval2023 competition, the reviewers, as well as all of those with whom I have had the pleasure to work during this and other related projects.

\bibliography{custom}
\bibliographystyle{acl_natbib}
\nocite{survay_Pretrained_Models}
\nocite{sexism}

\appendix

\section{Detailed Experimental Setup}
\begin{figure*}[t]
\centering
    \includegraphics[width=\textwidth, height=12cm]{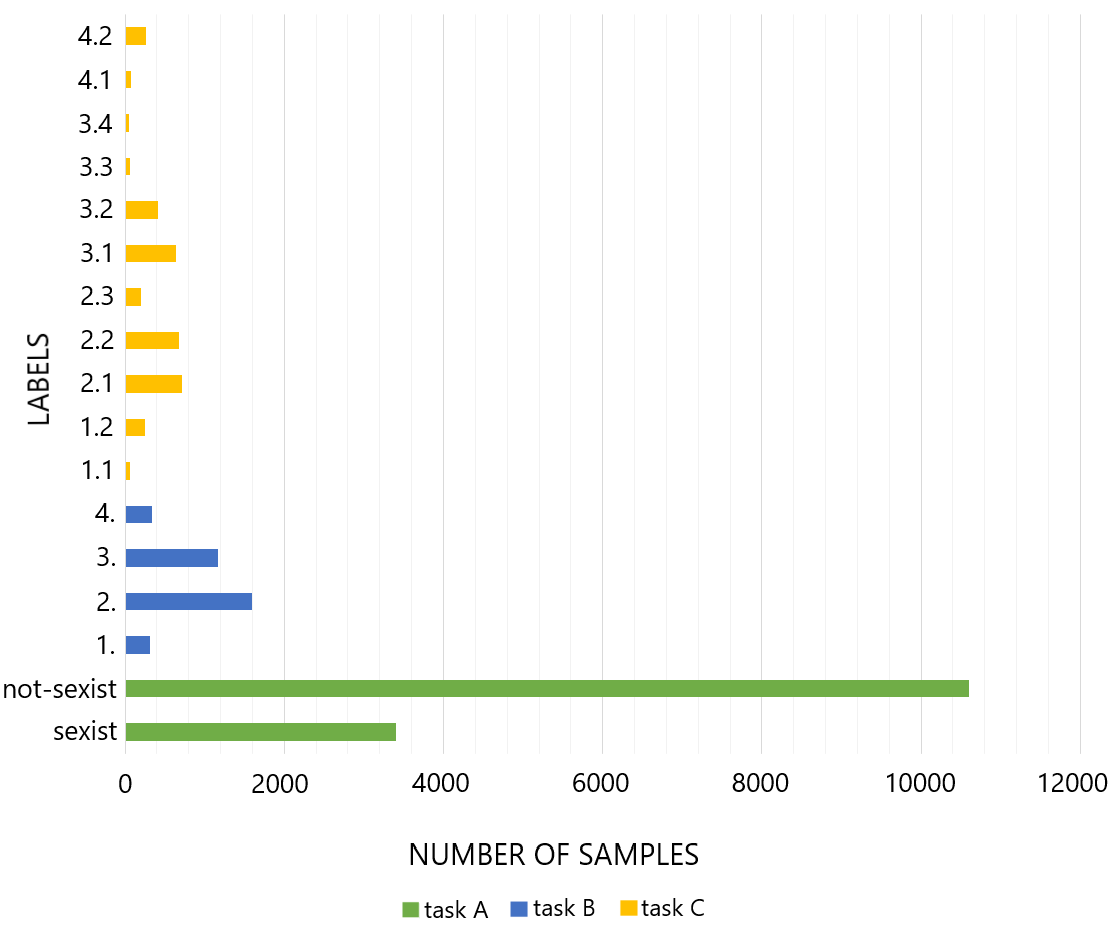}
    \caption{The train dataset distribution diagram of A, B, and C subtask classes.}
    \label{fig:label_dis}
\end{figure*}

\begin{figure*}[ht]
\centering
  \subcaptionbox*{Confusion matrix on subtask B}[0.48\linewidth]{%
    \includegraphics[width=\linewidth]{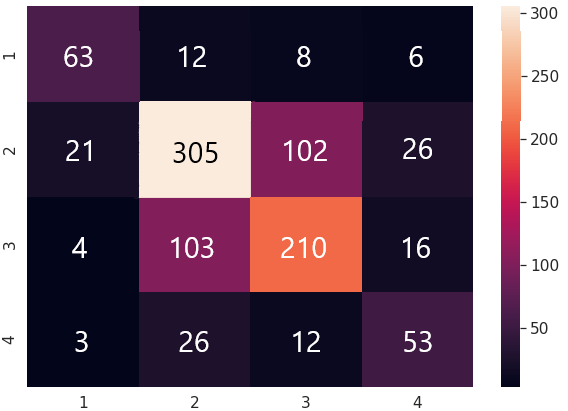}%
  }%
  \hfill
  \subcaptionbox*{Confusion matrix on subtask C}[0.48\linewidth]{%
    \includegraphics[width=\linewidth]{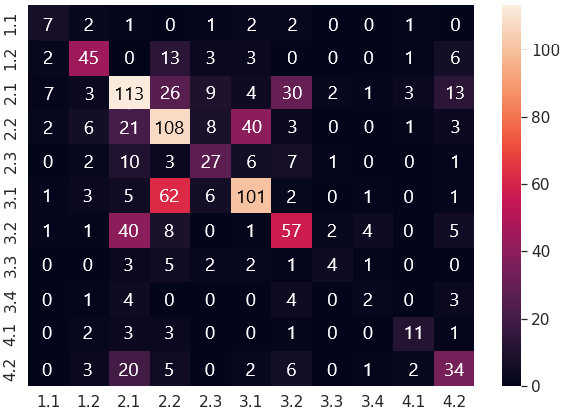}%
  }
    \caption{Confusion matrix of DeBERTa model with class-weight on subtasks B and C. These predictions were made on the dev sets. }
    \label{fig:confusion_BC}
\end{figure*}

The data is highly imbalanced across all three subtasks(see Figure \ref{fig:label_dis}). To deal with this problem, we used class-weight in the Loss function of our models. This method improved the F1-score by almost 2\% in the BERT model and for subtask A, and we observed very little improvement for subtasks B and C. But in other models such as DeBERTa, the class-weight was able to significantly improve the performance of the model on the test data in all three subtasks and increase the F1-score in most cases. Figure \ref{fig:confusion_BC} shows the confusion matrix of the B and C subtasks. These results come from the DeBERTa model, which uses class-weight in the loss function.
In this task, for each model, we first obtained the best hyperparameters using the dev dataset, then for the testing phase of the model, we combined the dev and training datasets and retrained the model, Then we evaluated the test dataset. This could increase the F1-score for all subtasks by an average of 1\%. 

\subsection{Task-Adaptive System}
In the task-adapted system, we trained the pre-trained BERT\(base\) model as an unsupervised for both Masked Language Model(MLM) and Next Sentence Prediction (NSP) tasks. To achieve this, we examined 2 million unlabeled datasets and filtered out those containing less than two sentences. Finally, about 900,000 data remained, with which we trained the model with a learning rate of \(2e-5\), batch size of 32, and in 2 epochs. Finally, we fine-tuned the model with labeled data.

\end{document}